\documentclass[cameraready]{Interspeech}
\title{Poly-InstructTTS: Learning In-the-Wild Expressive Speech Synthesis from Open-Ended Instructions}

\author{Junhui}{Zhang}
\author{Qianhui}{Xu}
\author{Qingxiang}{Guo}
\author{Dawei}{Yang}
\author{Ling}{Miao}
\author{Qiangqiang}{Wang}
\author{Yang}{Song}
\address{
    ZuoYeBang Technology, China
}

\email{\{zhangjunhui04, xuqianhui02, guoqingxiang, yangdawei02, miaoling, wangqiangqiang, songyang\}@zuoyebang.com}


\usepackage{comment}
\usepackage{cite}
\usepackage{multirow}
\usepackage{booktabs}
\usepackage{tabularx}
\usepackage{ragged2e}
\usepackage{subcaption}
\usepackage{graphicx}
\usepackage{array}
\usepackage{threeparttable}

\begin{document}

\maketitle

\begin{abstract}
While recent text-to-speech (TTS) models achieve high naturalness, controlling fine-grained expression via natural-language instructions remains challenging. We introduce \textbf{Poly-InstructTTS}, which learns expressive speech from open-ended instructions using in-the-wild audiovisual data. We build a scalable multi-modal pipeline to construct a 1,000-hour instruction-annotated corpus covering 1,000+ fine-grained emotions and styles. The framework uses a prompt-free GPT with attribute-based thinking tokens, followed by a flow-matching module that injects timbre from a reference audio. We also present a speaker fine-tuning procedure to transfer instruction control to specific speakers while preserving persona. We further extend InstructTTSEval with broader tasks. Experiments show that Poly-InstructTTS delivers strong performance in instruction adherence and expressiveness. Audio demos and the expanded testset are available on our project page\footnote{https://zhangjh915.github.io/PolyInstructTTS-demo/}.
\end{abstract}

\section{Introduction} \label{sec:intro}

For the past few years, GPT-based text-to-speech (TTS) models\cite{wang2023neural,wang2024maskgct,anastassiou2024seed,du2024cosyvoice,du2025cosyvoice,zhou2025indextts2,hu2026qwen3} have developed rapidly, leading to significant improvements in speech naturalness and expressiveness. The prevalent structure includes a GPT for discrete audio token prediction, a Flow-Matching (FM) acoustic model and a vocoder for waveform generation. These models perform well on zero-shot voice cloning, effectively capturing and reproducing the timbre and prosody from a short reference audio.

Beyond cloning, generating controllable speech using natural language instructions instead of predefined style tags has gained attention. However, the challenge becomes to acquire deep text-understanding capabilities to interpret complex instructions. Furthermore, mainstream zero-shot TTS systems usually require prompt audio as input to the GPT model, which may introduce conflicts between the prompt and target styles specified by the instruction, degrading the controllability.

Another challenge is the scarcity of expressive speech training data. Most open-source TTS datasets like LibriTTS\cite{zen2019libritts} and MLS\cite{pratap2020mls} are dominated by neutral reading tones, and some datasets include emotion labels\cite{adigwe2018emotional,zhou2022emotional} only cover a few major categories and lack natural language descriptions. Consequently, existing instruction-TTS systems struggle to handle subtle emotional shifts or extreme speaking styles (e.g., hesitation, screaming, whispering). 

To address the challenges, we propose \textbf{Poly-InstructTTS}, a TTS system capable of understanding diverse natural language instructions for expressive speech synthesis. The model is trained on dataset generated by a \textbf{poly-modal data pipeline} and the model is capable of \textbf{poly-style generation} for fine-grained emotions, styles, and paralinguistic behaviors with \textbf{poly-instruction adherence}. The contributions of this paper are summarized as follows:

\begin{itemize}
    \item We design a multi-modal data processing pipeline to generate diverse instruction-audio pairs from cinematic and television audiovisual media.
    \item We introduce a TTS framework with a prompt-free GPT guided by attribute-based thinking tokens and an FM module for timbre injection, together with a fine-tuning scheme to transfer controllability to specific speakers.
    \item We expand the InstructTTSEval\cite{huang2025instructttseval} testset to include more diverse evaluation tasks for the open-source community.
\end{itemize}

\section{Related Works} \label{sec:related_works}

\subsection{Instruction-Annotated Audio Datasets}

The development of instruction-controllable TTS heavily relies on high-quality instruction-text-audio pairs. Early efforts\cite{guo2023prompttts,leng2023prompttts} transitioned from simple concatenation of acoustic tags, such as gender, emotion, pitch, etc., to rule-based natural language generation. With the advent of Large Language Models (LLMs), more recent works\cite{vyas2023audiobox,lyth2024natural,jin2024speechcraft,chen2026flexivoice,ren2026ov} leverage LLMs to generate diverse, multi-attribute instruction texts.

However, most instruction datasets are derived from open-source datasets, recordings with limited emotions and styles such as audiobooks and podcasts, or synthesized speech from commercial APIs. Consequently, they are dominated by neutral, standardized reading tones and it's hard to capture paralinguistic phenomena or extreme emotional expressions in real-world conversations. In contrast, our work explicitly targets in-the-wild cinematic and television data to reach a higher ceiling.

\begin{figure*}[t]
    \centering
    % 第一张子图 (左侧)
    \begin{subfigure}[b]{0.48\textwidth}
        \centering
        % 保持长宽比，宽度设为子图宽度的100%
        \includegraphics[width=\linewidth]{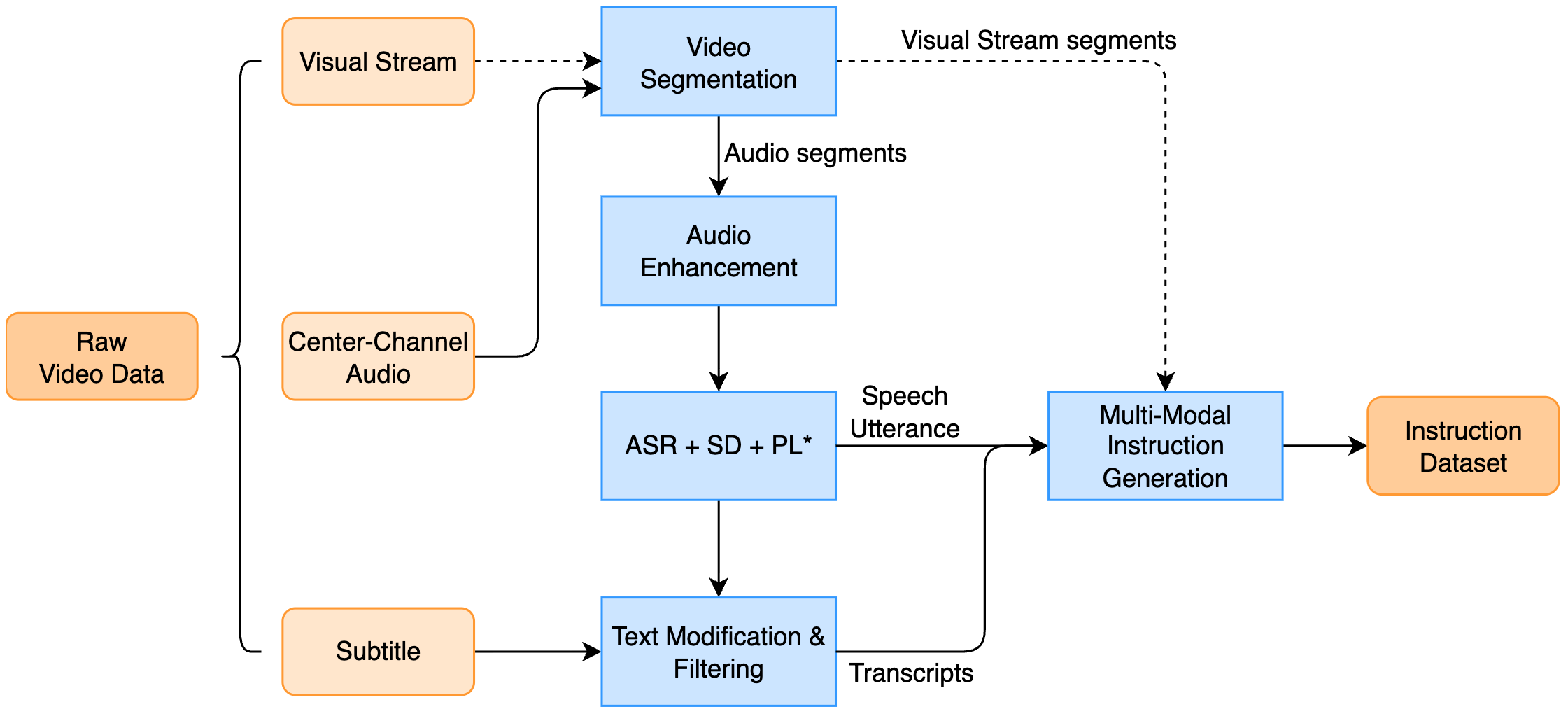}
        \caption{Proposed data pipeline. * ASR: Automatic Speech Recognition, SD: Speaker Diarization, PL: Paralinguistic Tagging via 11Labs API\cite{ElevenLabs-STT}}
        \label{fig:data_pipeline}
    \end{subfigure}
    \hfill % 这里的 hfill 很重要，它把两张图撑开，分别靠左和靠右对齐
    % 第二张子图 (右侧)
    \begin{subfigure}[b]{0.48\textwidth}
        \centering
        \includegraphics[width=\linewidth]{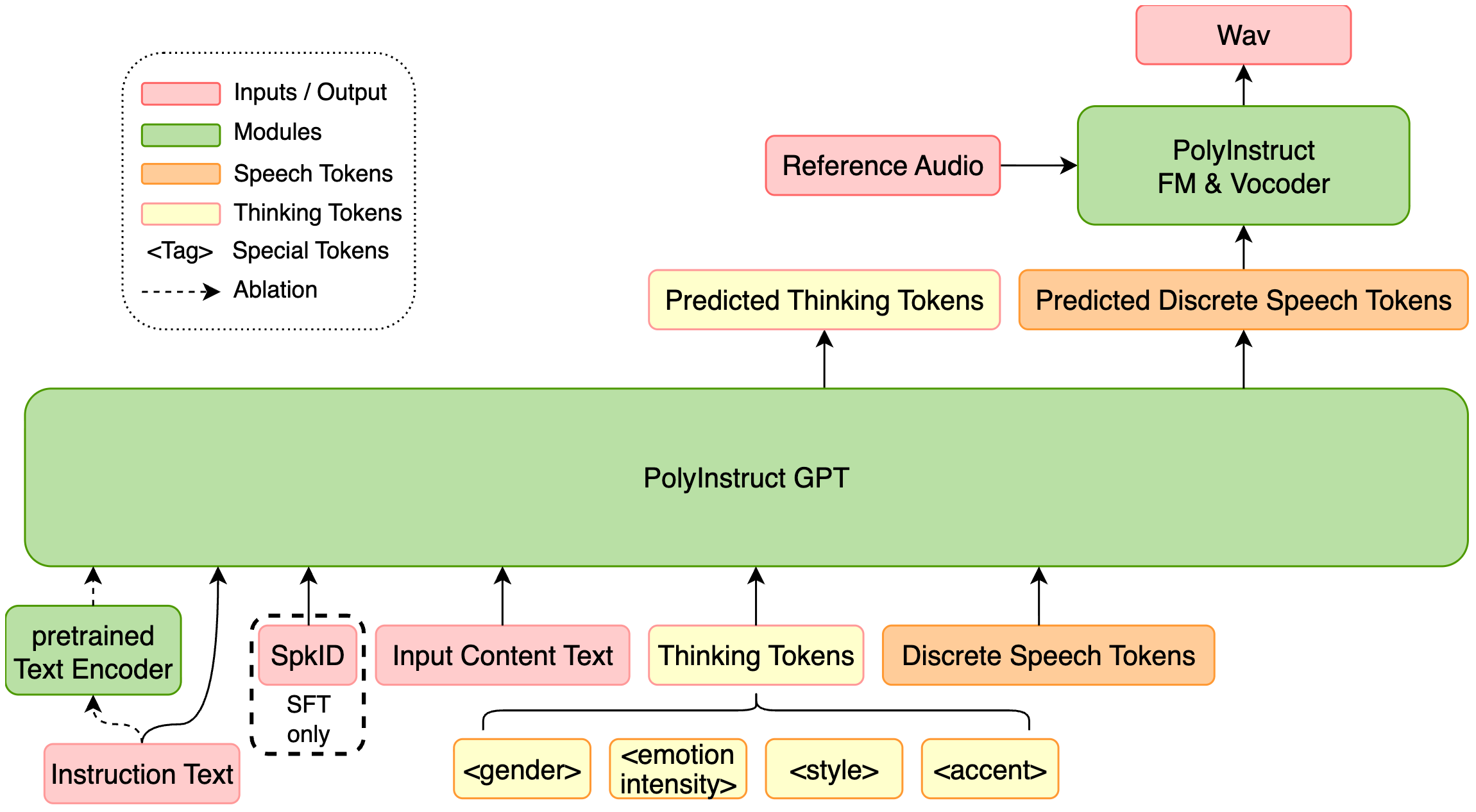}
        \caption{Proposed Poly-InstructTTS architecture.}
        \label{fig:model_arch}
    \end{subfigure}
    
    % 整个大图的标题和标签
    \caption{Overview of our proposed Poly-InstructTTS system. (a) illustrates the data pipeline for constructing the instruction dataset. (b) shows the model architecture of the Poly-InstructTTS based on GPT-FM framework.}
    \label{fig:system_overview}
\end{figure*}

\subsection{Instruction-Following TTS Systems}

Early instruction-based systems, such as InstructTTS\cite{yang2024instructtts} and PromptTTS\cite{guo2023prompttts,leng2023prompttts}, encode descriptions with traditional acoustic models, rather than leveraging GPT-based semantic modeling. Recent studies like FLEXIVOICE\cite{chen2026flexivoice} concatenates instruction text with content text as GPT input but still requires an audio prompt for inference. This dependency risks ``style leakage", where the prompt's acoustic properties may conflict with the intended instruction. EmoVoice\cite{yang2025emovoice} eliminates the prompt requirement but restricts its scope to major emotions. VoxInstruct\cite{zhou2024voxinstruct} introduces an MT5\cite{xue2021mt5} text encoder in GPT for diverse tasks with a codec language model. TextrolSpeech\cite{ji2024textrolspeech} converts text into phonemes as GPT inputs, which may destroy high-level semantic connections for instruction comprehension.

Related to our architecture, Qwen3-TTS\cite{hu2026qwen3} introduces the concept of ``thinking tokens" after the text input, yet the technical design and training pipeline remain undisclosed. OV-INSTRUCTTTS\cite{ren2026ov} adopts a GPT-FM framework and supervises training with explicit LLM‑generated ``thinking text". We argue that forcing the acoustic model to predict complex, free-form thinking text introduces unnecessary optimization difficulties. In our setting, we observe that such a strategy can harm stability. In contrast, our system employs attribute-based thinking tokens, effectively steering the style generation without degrading the model's intrinsic semantic understanding.

\section{Method} \label{sec:methods}

\subsection{Dataset Construction Pipeline}

\subsubsection{Data Source} \label{subsec:data}

To capture the rich emotions and paralinguistic behaviors rarely found in traditional reading-style corpora, we compiled an in-the-wild 1,000-hour dataset from cinematic and television audiovisual media. These sources contain expressive, context-driven conversational interactions, abundant emotions, diverse paralinguistic behaviors, and various accents. The media files are collected from publicly accessible sources and processed strictly for non-commercial, academic research purposes. We construct a data pipeline in Figure\ref{fig:data_pipeline} to generate our instruction dataset as detailed described in Section \ref{subsec:alignment} and \ref{subsec:gemini}.

\textit{Due to copyright restrictions of commercial cinematic media, the raw dataset cannot be publicly distributed. However, we detail the pipeline and open-source the LLM prompts for researchers to reproduce on their own corpora.}

\subsubsection{Audio-Text Segmentation \& Alignment} \label{subsec:alignment}

To obtain clean single-speaker utterances, we propose a cascaded data processing pipeline as Figure\ref{fig:data_pipeline}. First, we segment the videos into few-minute clips based on long silences detected by subtitle timestamps and a Voice Activity Detection (VAD) tool\cite{Silero_VAD}. To eliminate post-production sound effects and background music, we extract the center channel audio track from MKV files and denoise it with Demucs\cite{defossez2019demucs} and ClearerVoice\cite{zhao2025clearervoice} sequentially. Then we use a commercial service, ElevenLabs’ Speech-to-Text API\cite{ElevenLabs-STT}, including automatic speech recognition (ASR), speaker diarization (SD) and paralinguistic tagging (PL) to generate single-speaker sentence-level transcription with fine-grained paralinguistic tags. For example, \textit{Anna \texttt{<breathing>} please, \texttt{<whisper>} someone is there... \texttt{</whisper>} Just kidding \texttt{<laughter>}!} To ensure text accuracy, we employ a rule-based fuzzy matching strategy between the ASR outputs and subtitles. Sentences with a low Word Error Rate (WER) against the subtitles are replaced with the definitive subtitle text, while segments with a high WER are discarded as misrecognitions of background speech.

\subsubsection{Multi-Modal Instruction Generation} \label{subsec:gemini}

With the text-audio pairs prepared, we utilize a commercial service of multi-modal LLM, Gemini 2.5 Pro API\cite{Google-Gemini-API}, to generate natural language instructions via a three-stage pipeline. The prompts for all stages are available in the demo page.

\textbf{Stage 1 - Content Summary:} LLM integrates down-sampled video (720p) with audio to extract the narrative structure, character relationships, and global contextual background.

\textbf{Stage 2 - Transcript Analysis:} Guided by the Stage-1 summary, the sentence-level transcripts and the video clip, the LLM annotates each sentence with multi-dimensional attributes including gender, accent, emotion and speaking style. The multi-modal input is critical for capturing subtle, fine‑grained emotions that are difficult to infer from audio‑only input.

\textbf{Stage 3 - Instruction Generation:} Based on the global context and segment-level attributes, the LLM generates a vivid, diverse natural language instruction for each utterance.

\subsection{Poly-InstructTTS Framework}

As illustrated in Figure\ref{fig:model_arch}, Poly-InstructTTS follows a GPT-FM architecture, including an auto-regressive GPT to translate instruction and content texts into discrete speech tokens, an FM module to convert these tokens into mel-spectrograms, and a HiFi-Net\cite{li2023hiftnet} to reconstruct the audio waveform. From an architectural perspective, the system is designed to route prosody and style to the GPT while injecting timbre only at the acoustic module. The GPT is prompt-audio-free and an FM model from CosyVoice3\cite{du2025cosyvoice} injects timbre conditioned on the reference audio. In this way, the GPT learns to map instructions to content and prosody, and this setup mitigates style leakage between reference and instruction while preserving speaker similarity.

The GPT first undergoes a pre-training stage on large-scale speech corpora of Emilia\cite{he2024emilia} and MLS\cite{pratap2020mls} to acquire foundational TTS capabilities. For the post-training stage on our instruction dataset, each sample is serialized as \texttt{<Instruct> <Text> <Thinking Tokens> <Speech Tokens>}. For example: \texttt{<instruct> He lets out a scream, the words drawn slowly with a desperate plea filled with fear. </instruct> <text> Oh God, please no! </text> <think> <male> <high> <shouting> </think> <SpeechTokens>}. Instead of using LLM-generated  thinking text tokens as in OV-InstructTTS\cite{ren2026ov}, we adopt a compact set of attribute‑based thinking tokens to bridge instructions and acoustic tokens. We argue that predicting complex LLM text introduces unnecessary optimization burdens and risks catastrophic forgetting of the content-understanding abilities acquired during pre-training. The thinking token set includes \textit{gender}, \textit{emotion intensity}, \textit{style}, and \textit{accent}. Style and accent are selected as the top labels in the training corpus that occur 100+ times and all attribute-based tokens are treated as special tokens for GPT. We deliberately exclude other attributes like pitch and emotion from the thinking tokens, so that the GPT remains responsible for realizing prosody from the instruction and the thinking tokens. Practically, the \textit{gender} token can reduce the pitch–timbre mismatches in FM (e.g., high‑pitch prosody on a low‑pitch reference), improving stability and speaker similarity.

\subsection{Instruction-Conditioned Speaker Fine-Tuning}

While the base Poly-InstructTTS model is capable of generating zero-shot style with arbitrary timbres, real-world applications often require endowing a specific speaker with instruction-following capabilities while maintaining their intrinsic persona. For this scenario, we propose an Instruction-Conditioned Speaker Fine-Tuning (SFT) scheme to transfer the expressive variance from our cinematic dataset to target speakers. The training format of the GPT remains identical to the base stage, but we prepend a Speaker ID tag to the content text. For the cinematic data, the tag is set to \texttt{<Unknown>}, whereas SFT data utilizes specific IDs such as \texttt{<spk\_01>}. During inference, simply conditioning the prompt with the target Speaker ID allows the GPT model to seamlessly blend the learned instruction controllability with the fine-tuned speaker characteristics.

\section{Experiments} \label{sec:exp}

\subsection{Experimental Setup}

\subsubsection{Training Dataset \& Splits} \label{subsec:data_split}

Following the proposed pipeline, we processed a raw corpus of approximately 2,500 hours, yielding a highly expressive dataset of 1,000 hours with over 1.1M utterances, and covers over 200 accents, 800 fine-grained emotions and 400 stylistic variations annotated by the LLM. For attribute-based thinking tokens\footnote{full category lists are provided in the supplementary material.}, we retain \textit{style} and \textit{accent} labels with 200+ occurrences. The instruction–audio match rate exceeds 95\% according to human ratings. We randomly split the corpus into 99\% training and 1\% validation. For the subsequent Instruction-SFT task, we additionally curate 200 hours of speech from 10 distinct speakers.

\subsubsection{Evaluation Testset}

Our primary evaluation is based on the InstructTTSEval\cite{huang2025instructttseval} benchmark. However, this original benchmark under-represents certain scenarios such as diverse accents, subtle/extreme emotions, and non-mainstream styles. To address these gaps, we introduce an expanded testset of 200 samples with the same format as the complementary of the base testset. We evaluate on both the base and the expanded test sets. Table\ref{tab:data_comparison} describes the comparison between these testsets, and the expanded testset is available in our project page.

\begin{table}[t]
    \centering
    \footnotesize
    \caption{Comparison of attribute distribution between the base InstructTTSEval and the proposed Expanded Testset.}
    \label{tab:data_comparison}
    \begin{tabularx}{\linewidth}{@{}l>{\RaggedRight}X>{\RaggedRight\arraybackslash}X@{}}
        \toprule
        \textbf{Dimension} & \textbf{InstructTTSEval} & \textbf{Expanded Testset} \\
        \midrule
        \textbf{Accent} & \textbf{Native Standard} \newline (US/UK) & \textbf{Accents \& Dialects} \newline (Indian, Hispanic...) \\
        \addlinespace[0.4em]
        \textbf{Emotion} & \textbf{Common Intensity} \newline (Angry, Terrified, Complaint...) & \textbf{Extreme Variants} (Scream, Sarcastic...) \newline \textbf{Low Intensity} (Bored, Relieved...) \\
        \addlinespace[0.4em]
        \textbf{Style} & \textbf{Mainstream Styles} \newline (Broadcast, Narration, Drama...) & \textbf{Diverse Personas} \newline (Robotic, Meditation, ASMR...) \\
        \addlinespace[0.4em]
        \textbf{Fluency} & \textbf{Fluent} \newline (Read speech) & \textbf{Spontaneous} \newline (Hesitation, Filler words, Stuttering) \\
        \addlinespace[0.4em]
        \textbf{Acoustic} & \textbf{Standard Quality} \newline (Clean) & \textbf{In-the-wild} \newline (Telephone, Echo...) \\
        \bottomrule
    \end{tabularx}
\end{table}

\subsubsection{Evaluation Metrics}

For objective assessment, we measure the system stability using Word Error Rate (WER) and the instruction-following capabilities using the metrics defined in InstructTTSEval: Acoustic-Parameter Specification (APS) for basic acoustic instruction following, Descriptive-Style Directive (DSD) for style assessment, and Role-Play (RP) for evaluation of role play ability.

Given that LLM-based objective evaluators often struggle to reliably assess subtle paralinguistic nuances and extreme human emotions, subjective human evaluation is also conducted using mean opinion score (MOS) tests with 20 native listeners. The evaluation encompasses two dimensions on a 5-point scale: \textbf{I-MOS} (instruction-following accuracy) and \textbf{N-MOS} (speech naturalness and stability).

\newcolumntype{C}{>{\centering\arraybackslash}p{0.85cm}} 

\begin{table*}[t]
\centering
\begin{threeparttable}
\caption{Comparison of the baselines and proposed systems with ablations on the performance of InstructTTSEval \textbf{Base} and \textbf{Expanded} testsets. The systems include closed-source APIs, open-source models and the proposed system with ablations.}
\label{tab:overall_results}
\small
\setlength{\tabcolsep}{3.5pt}
\begin{tabular}{l c | CCCCC | CCCCC}
\toprule
\multirow{2}{*}{\textbf{System}} & \textbf{General} & \multicolumn{5}{c|}{\textbf{InstructTTSEval (Base)}} & \multicolumn{5}{c}{\textbf{InstructTTSEval (Expanded)}} \\
 & \scriptsize{\textbf{WER}$\downarrow$} & \scriptsize{\textbf{APS}$\uparrow$} & \scriptsize{\textbf{DSD}$\uparrow$} & \scriptsize{\textbf{RP}$\uparrow$} & \scriptsize{\textbf{IMOS\tnote{*}}$\uparrow$} & \scriptsize{\textbf{NMOS\tnote{*}}$\uparrow$} & \scriptsize{\textbf{APS}$\uparrow$} & \scriptsize{\textbf{DSD}$\uparrow$} & \scriptsize{\textbf{RP}$\uparrow$} & \scriptsize{\textbf{IMOS\tnote{*}}$\uparrow$} & \scriptsize{\textbf{NMOS\tnote{*}}$\uparrow$} \\
\midrule
% \multicolumn{12}{l}{\textit{\textbf{Closed-Source Commercial APIs}}} \\
Gemini-Pro TTS   & 2.92 & 87.6 & 86.0 & 67.2 & 3.70 & \textbf{4.28} & 72.5 & 90.9 & 86.4 & 3.82 & 3.95 \\
Gemini-Flash TTS & 2.75 & \textbf{92.3} & \textbf{93.8} & 80.1 & 3.71 & 4.26 & 77.4 & \textbf{92.9} & 87.0 & 3.87 & 3.57 \\
GPT-4o-mini TTS  & 2.96 & 76.4 & 74.3 & 54.8 & 2.66 & 3.95 & \textbf{78.0} & 84.9 & 91.5 & 2.80 & 3.64 \\
\midrule
% \multicolumn{12}{l}{\textit{\textbf{Open-Source Baselines}}} \\
Qwen3-TTS-12Hz-1.7B-VD~\cite{hu2026qwen3} & \textbf{2.09} & 82.9 & 82.4 & 68.4 & 3.76 & 4.16 & 73.0 & 92.5 & 88.0 & \textbf{4.12} & \textbf{4.24} \\
OV-InstructTTS~\cite{ren2026ov}        & 17.47 & 78.3 & 77.8 & 61.3 & 3.05 & 3.13 & 61.0 & 85.9 & 70.0 & 3.20 & 3.30 \\
Mimo-Audio-7B-Instruct~\cite{zhang2025mimo} & 5.68 & 80.6 & 77.6 & 59.5 & 3.33 & 2.98 & 66.5 & 91.5 & 85.4 & 3.42 & 3.11 \\
VoxInstruct~\cite{zhou2024voxinstruct} & 9.44 & 54.9 & 57.0 & 39.3 & 2.59 & 3.22 & 55.0 & 88.4 & 66.9 & 2.71 & 3.60 \\
Parler-tts-large~\cite{lyth2024natural} & 15.67 & 60.0 & 45.9 & 31.2 & 2.47 & 2.87 & 50.0 & 65.9 & 68.5 & 2.74 & 3.39 \\
PromptTTS~\cite{leng2023prompttts}     & 2.50 & 64.3 & 47.2 & 31.4 & 2.34 & 3.43 & 69.5 & 67.0 & 68.5 & 2.49 & 3.52 \\
PromptStyle~\cite{guo2023prompttts}    & 11.39& 57.4 & 46.4 & 30.9 & 2.22 & 2.83 & 69.0 & 63.8 & 69.5 & 2.51 & 2.98 \\
\midrule
% \multicolumn{12}{l}{\textit{\textbf{Proposed System}}} \\
\textbf{Poly-InstructTTS}            & 4.42 & 91.1 & 79.2 & \textbf{88.7} & \textbf{3.81} & 3.88 & 77.1 & 84.5 & 90.1 & 3.96 & 3.36 \\
\hspace{2mm} \textit{without thinking tokens} & 4.53 & 88.9 & 80.7 & 86.0 & 3.63 & 3.69 & 74.4 & 79.7 & 84.0 & 3.58 & 3.42 \\
\hspace{2mm} \textit{with FlanT5\cite{longpre2023flan} text encoder}  & 4.84 & 89.7 & 81.3 & 85.0 & 3.74 & 3.76 & 75.0 & 83.0 & 92.0 & 3.54 & 3.27 \\
\hspace{2mm} \textit{with Instructor\cite{ni2022large} text encoder}  & 5.75 & 88.0 & 81.1 & 85.1 & 3.77 & 3.68 & 77.8 & 83.5 & 93.5 & 3.38 & 3.09 \\
\hspace{2mm} \textit{with GTR-base\cite{su2023one} text encoder}  & 4.91 & 89.4 & 81.1 & 86.3 & 3.79 & 3.92 & 72.0 & 85.0 & \textbf{95.0} & 3.27 & 3.38 \\
\bottomrule
\end{tabular}
\begin{tablenotes}[flushleft]
\footnotesize
\item[*] The 95\% confidence intervals (95\% CI) for all reported MOS values are $\leqslant\pm$ 0.10. Only the mean scores are presented above for brevity.
\end{tablenotes}
\end{threeparttable}
\end{table*}

\subsubsection{Implementation Details}

The GPT backbone is initialized from Qwen2.5-0.5B-Instruct\cite{qwen2.5}. Speech tokens are extracted using the CosyVoice2\cite{du2025cosyvoice} speech tokenizer at 25Hz. The model is trained on 8 NVIDIA A800 GPUs for 30 epochs and optimized using the AdamW optimizer\cite{kingma2014adam} with a learning rate of 1e-4 and a batch duration of 300s. During inference, the FM and vocoder are consistent with CosyVoice3\cite{du2025cosyvoice}, and the reference audios are randomly chosen from LibriTTS\cite{zen2019libritts}. the overall system achieves a Real-Time Factor (RTF) of roughly 0.06 on a single RTX 4090 GPU.

\subsection{Evaluation Results}

We compare Poly-InstructTTS against the instruction-following TTS systems including open-source baselines and closed source systems. As shown in Tabel\ref{tab:overall_results}, Poly-InstructTTS achieves consistently high APS/DSD/RP scores, with the largest gains on RP, likely benefiting from our in-the-wild training data and thinking-token design. The moderate WER of our system may result from the challenging acoustic dynamics and label noise in the dataset compared to clean speech corpora. Subjectively, our system achieves top‑1 I-MOS on the base testset and top‑2 on the expanded set, demonstrating leading instruction adherence and expressiveness. The optional comments in subjective tests indicate that our system can handle diverse styles and scenarios and produce expressive speech with fine-grained emotions.

Limitations and trade-offs: evaluations also reveal weaknesses under challenging acoustic conditions (e.g., noise, echo), which appear more related to the FM than to the GPT instruction pathway. During post-training, we observe a stability–expressiveness trade-off that as epochs increase, APS/DSD/RP steadily improve while WER rises. We hypothesize that training on extreme emotions and paralinguistic behaviors weakens monotonic text–token alignment. These limitations remain a future work for our system.

\subsection{Ablation Studies}

To validate the core architectural designs, we conduct ablations in Table\ref{tab:overall_results}. First, removing the attribute‑based thinking tokens lowers subjective scores, indicating they provide a useful prior that helps align instructions with acoustic tokens. These tokens complement the free‑form instruction text, leaving the GPT to achieve better expressiveness compared to direct generation. We also evaluate optional pretrained instruction encoders, including FlanT5\cite{longpre2023flan}, instructor\cite{ni2022large} and GTR-base\cite{su2023one}, by conditioning the GPT on their frozen hidden representations of the instruction text (dashed lines in Figure\ref{fig:model_arch}). Across settings, these variants does not yield positive gains versus the encoder‑free default. This suggests that feeding raw instruction text directly to the GPT is sufficient for generating styled speech under our training setup.

\subsection{Instruction-Conditioned SFT Analysis}

Finally, we evaluate the model's performance on the speaker fine-tuning to endow specific speakers with instruction control while preserving their personas. We compared two systems: (1) the proposed SFT paradigm, and (2) the base Poly‑InstructTTS with only an FM‑stage timbre injection, both of which use the same timbre reference for FM.

We conduct a subjective study on 10 fine‑tuned speakers in Table\ref{tab:sft_comparison}. In addition to I‑MOS and N‑MOS, we report P‑MOS to indicate how closely the speech aligns with the speaker’s persona. Results indicate that the SFT variant attains higher P‑MOS, whereas the base setup yields higher I‑MOS. This suggests that SFT learns a speaker‑specific mapping from instructions to a suitable stylistic range, and is preferable despite the trade‑off in instruction adherence. In addition, SFT achieves a higher N‑MOS, likely reflecting the higher quality and consistency of the SFT corpus.

\begin{table}[th]
\centering
\caption{Subjective evaluation comparing the proposed Instruction-Conditioned SFT versus the base method.}
\label{tab:sft_comparison}
\resizebox{\columnwidth}{!}{%
\begin{tabular}{l ccc}
\toprule
\textbf{Method} & \textbf{I-MOS $\uparrow$} & \textbf{P-MOS $\uparrow$} & \textbf{N-MOS $\uparrow$} \\
\midrule
\textbf{Proposed SFT method} & 3.30 $\pm$ 0.08 & \textbf{3.91} $\pm$ 0.07 & \textbf{4.02} $\pm$ 0.09 \\
Base method & \textbf{3.60} $\pm$ 0.08 & 3.62 $\pm$ 0.08 & 3.76 $\pm$ 0.10 \\
\bottomrule
\end{tabular}%
}
\end{table}

\section{Conclusion} \label{sec:conclusion}

In this paper, we introduce Poly-InstructTTS, a highly expressive, instruction-following TTS framework that bridges semantic understanding and complex acoustic realization. To address the data scarcity, we build a multi-modal pipeline to construct a 1,000-hour in-the-wild cinematic corpus covering 1,000+ fine-grained emotions and styles. Architecturally, we adopt a GPT-FM framework with a prompt-free GPT with attribute-based thinking tokens to follow the open-ended natural language instructions. We also extend the InstructTTSEval testset. Experiments show that Poly-InstructTTS achieves strong performance in both objective and subjective metrics, comparable to the best baseline systems. Furthermore, our instruction-conditioned SFT strategy effectively endows target speakers with instruction controllability while preserving speaker personas.

Future work will focus on better balancing the trade-off between expressiveness and stability. We will also improve the FM module to cope with more complex conditions. Finally, we plan to move beyond reference-conditioned timbre and enable text-only, reference-free voice generation, preserving controllability while removing the need for a timbre reference audio.

\ifcameraready

    \section{Acknowledgments}

        The authors would like to express their sincere gratitude to the 20 human participants from ZuoYeBang Technology for their dedicated effort in conducting the subjective Mean Opinion Score (MOS) evaluations. Their invaluable feedback significantly contributed to the rigorous assessment of our proposed system.

\else
    
\fi

\section{Generative AI Use Disclosure}

    We used OpenAI's GPT‑5 and Google’s Gemini 3.0 Pro solely to polish the English language and improve readability during manuscript preparation. The tools were not used to generate scientific ideas, methods or experimental results. All technical content, analyses, claims, and conclusions are authored by the human authors. All AI‑assisted edits were reviewed and verified, and the authors take full responsibility for the final text. No confidential data or personally identifiable information were shared with the AI services.

\bibliographystyle{IEEEtran}
\bibliography{mybib}

\end{document}